\pdfoutput=1
\documentclass[10pt,twocolumn,letterpaper]{article}

\usepackage{cvpr}
\usepackage{times}
\usepackage{epsfig}
\usepackage{graphicx}
\usepackage{amsmath}
\usepackage{amssymb}

\usepackage{placeins}
\usepackage{color}
\usepackage{tabularx}
\usepackage{fixltx2e}
\usepackage{microtype}
\usepackage{makecell}

\usepackage{multirow}

\newenvironment{packed_item}{
\begin{itemize}
\vspace{-8pt}
  \setlength{\itemsep}{0pt}
  \setlength{\parskip}{0pt}
  \setlength{\parsep}{0pt}
  \setlength{\topsep}{-10pt}
  \setlength{\partopsep}{0pt}
}{\end{itemize}}

\usepackage[breaklinks=true,bookmarks=false]{hyperref}

\cvprfinalcopy 

\def\cvprPaperID598 
\def\httilde{\mbox{\tt\raisebox{-.5ex}{\symbol{126}}}}

\newcommand\wid{.31\textwidth}
\newcommand\widh{.15\textwidth}

\newcommand{\zbontar}
{
	\v{Z}bontar and  LeCun
}

\newcommand{\pmpar}[1]
{
\vspace{-1.0\baselineskip}
\paragraph{\textbf{#1}}
}

\begin{document}

\title{CBMV: A Coalesced Bidirectional Matching Volume for Disparity Estimation}

\author{Konstantinos Batsos \\
{\tt\small kbatsos@stevens.edu} 
\and
Changjiang Cai\\
{\tt\small ccai1@stevens.edu}
\and
Philippos Mordohai\\
{\tt\small mordohai@cs.stevens.edu }\\
}
\affiliation{Stevens Institute of Technology}

\maketitle

\begin{abstract}

Recently, there has been a paradigm shift in stereo matching with learning-based methods achieving the best results on all popular benchmarks. The success of these methods is due to the availability of training data with ground truth; training learning-based systems on these datasets has allowed them to surpass the accuracy of conventional approaches based on heuristics and assumptions. Many of these assumptions, however, had been validated extensively and hold for the majority of possible inputs. In this paper, we generate a matching volume leveraging both data with ground truth and conventional wisdom. We accomplish this by coalescing diverse evidence from a bidirectional matching process via random forest classifiers. We show that the resulting matching volume estimation method achieves similar accuracy to purely data-driven alternatives on benchmarks and that it  generalizes to unseen data much better. In fact, the results we submitted to the KITTI and ETH3D benchmarks were generated using a classifier trained on the Middlebury 2014 dataset.  
\end{abstract}


\section{Introduction}\label{sec:intro}
The most important recent development in stereo matching is the prevalent use of machine learning techniques that have led to dramatic improvements in accuracy by taking advantage of datasets with ground truth. Methods based on learning are effective because they replace assumptions and hand-crafted rules with data-driven, optimized decision rules and predictions. Classifiers are used to contribute in various stages of the disparity estimation process; several authors have trained classifiers to predict whether two image patches are likely to match \cite{chen15deep_embed,luo_urtasun16,park_lwcnn16,shaked17,ye_ieee17,zagoruyko15,zbontar15,zbontar16}, while others have used classifiers to replace hand-crafted rules in other stages of the process \cite{gidaris17,park15,poggi16_3dv_fusion,poggi16_3dv,seki16,seki17,spyr_3DV2015,spyropoulos_ijcv16}.
A quick inspection of the most active binocular stereo benchmarks \cite{geiger13,Menze2015CVPR,scharstein14} reveals that learning the matching function, in particular the work of \zbontar \cite{zbontar16}, has been the primary enabling technology behind the majority of the top-ranked algorithms.

\begin{figure}[b]
\begin{center}
\begin{tabular}{cc}
\includegraphics[width=.21\textwidth]{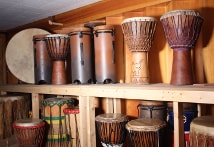} & \includegraphics[width=.21\textwidth]{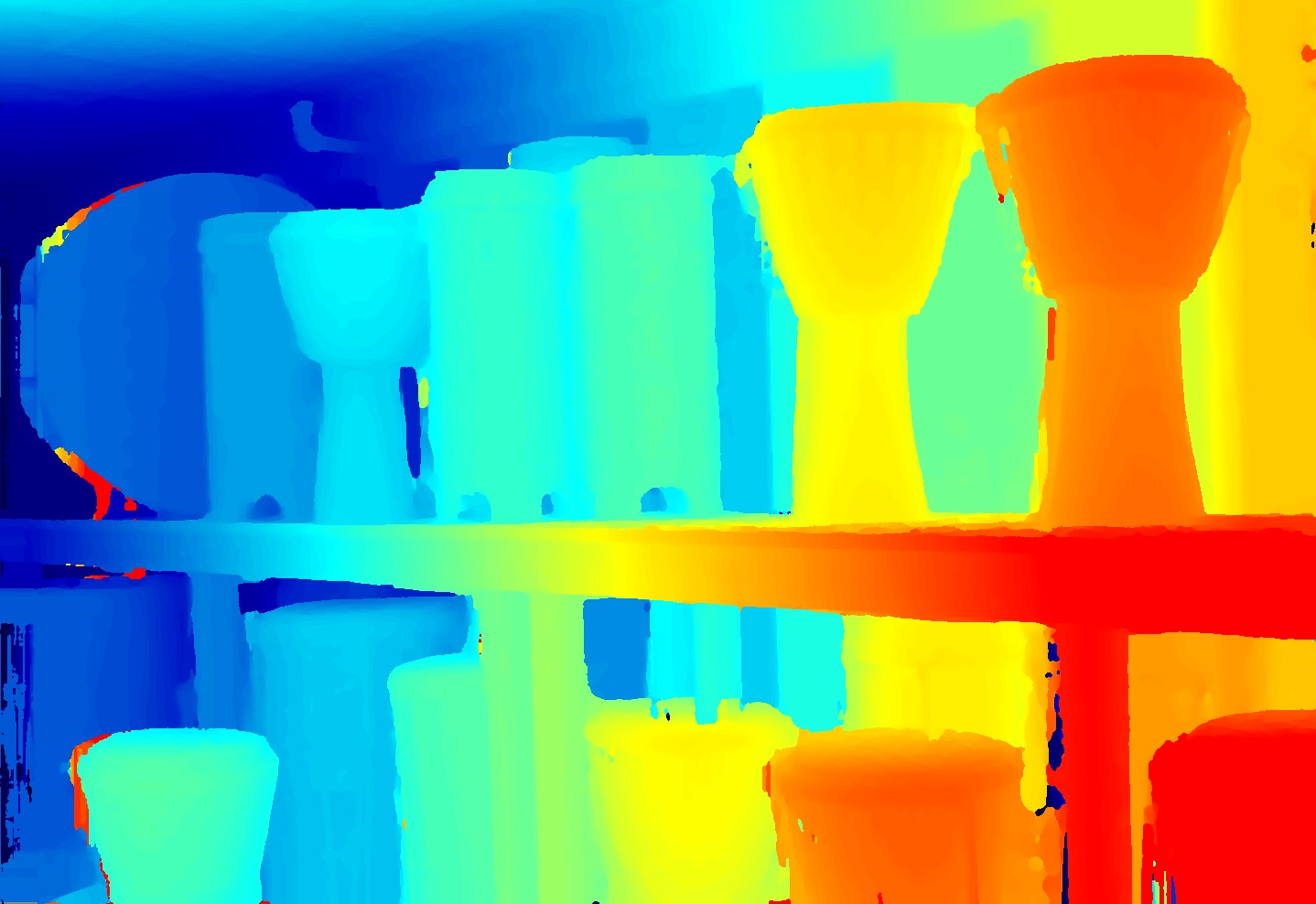} \\
Djembe & CBMV disparity map \\
\end{tabular}
\end{center}
\vspace{-5pt}
\caption{The left view of Djembe stereo dataset \cite{scharstein14} along with the disparity map computed by CBMV }
\label{fig:MVS_piano}
\end{figure}

In addition to the above approaches that propose learning-based components integrated into the stereo matching pipeline, there are a few deep learning architectures that allow end-to-end training \cite{kendall17,knobelreiter17,mayer16,pang17}. While end-to-end training has the advantage that it avoids suboptimal configurations, which often occur when intermediate objectives are optimized separately from final disparity map accuracy, its downside is that these methods tend to over-specialize in the training domain. As anecdotal evidence for this statement we provide the observation that very few results from end-to-end architectures have been submitted to the Middlebury 2014 benchmark \cite{scharstein14}, which contains images of various resolutions and disparity ranges.
Specialization is a desirable property in many applications, such as autonomous driving. In this paper, we aim to create a \emph{general approach that can be effective on a broad range of input imagery.}


In contrast to end-to-end architectures, deep learning methods that learn the matching likelihood of image patches have shown better generalization properties, but they cannot be trained in an end-to-end manner. For example, the MC-CNN method of \zbontar \cite{zbontar15,zbontar16} uses a Siamese CNN to estimate the matching volume, that contains the matching likelihood/cost for each allowable disparity of every pixel, and conventional steps to optimize the volume and extract the final disparity map. Their pipeline resembles the last three steps of the conventional pipeline according to Scharstein and Szeliski \cite{scharstein02}: cost aggregation, disparity optimization and disparity refinement. MC-CNN has been widely adopted as the cost function by a number of authors who have presented state-of-the-art results \cite{barron16,drouyer17,gidaris17,kim16ntde,li_3dmst17,seki16,seki17,taniai17}.

Our goal is similar to MC-CNN, since we also aim to estimate a matching volume that can be used as input to various optimization algorithms enabling them to produce highly accurate disparity maps. Instead of taking an extremely data-driven approach, in which the stereo matcher is only provided with two image patches and a label specifying whether they match, we design our matching volume estimator with an emphasis on \emph{robustness} and \emph{invariance}.

To improve the generalizability of our approach, we design it to be invariant to common variations of the input images. (It may be possible to achieve invariance by applying data augmentation techniques to the training set, but then the designer would have to specify the variations manually.)
Most conventional matching functions provide some form of invariance to specific transformations of the input. For instance, normalized cross-correlation (NCC) is invariant to affine intensity transformations, while the census transform \cite{zabih94} is invariant to transformations that preserve the ordering of intensities in the matching window.
These matching functions are known to fail quite often, but their failures can be predicted via the use of confidence measures \cite{hu12,poggi17_iccv}. More importantly, these failures are mitigated by combining a diverse set of matching functions.

In addition to using four matching functions in the current implementation of our approach, we compute two measures of confidence for each matching function and each matching direction: left-to-right and right-to-left. The matching cost between a pixel $p_L$ in the left image and a pixel $p_R$ in the right image is the same regardless of the matching direction. The ambiguity, and thus the confidence, of the correspondence, however, may differ with respect to the matching direction. A disparity assignment, that joins a pixel in the left image with one in the right image, must compete with other possible disparity assignments in both the left and the right epipolar line. Unlike most previous work \cite{haeusler13,park15,spyropoulos_ijcv16}, we measure the degree of competition (ambiguity) in both directions. The resulting matching and confidence data are coalesced by a random forest (RF) classifier \cite{breiman01} that estimates the probability of correctness of each disparity. Hence, we name our method \emph{Coalesced Bidirectional Matching Volume (CBMV)}. Figure \ref{fig:MVS_piano} shows an example disparity map estimated by our algorithm.

Throughout the paper, we compare our approach to MC-CNN  \cite{zbontar16}.
In order to make the comparison straightforward, we apply their optimization and post-processing pipeline on the CBMV volume.
Our experiments show that CBMV generalizes to data from domains that differ substantially from the training domain. We believe that the reason for this is that \emph{our approach learns to reason on relationships in the matching volume without being affected by image appearance,} which it never observes directly.

The contributions of this paper are:
\begin{packed_item}
\item A novel method for computing the matching volume for stereo that benefits from the combination of multiple matching functions and confidence estimates in both matching directions.
\item Competitive results with the fast MC-CNN architecture, which is the most widely adopted deep architecture for patch matching.
\item An improved capability to generalize to inputs from unseen domains much better than deep learning approaches, such as MC-CNN.
\end{packed_item}

\section{Related Work}\label{sec:related}
For a general survey of binocular stereo methods we refer readers to \cite{scharstein02}. In this section, we review learning-based methods in which learning is directly relevant to disparity estimation.
We consider methods that learn hyper-parameters of the stereo algorithm \cite{pal12,trinh09,yamaguchi12,zhangSeitz07} out of scope. We classify the methods below according to the primary problem they address: determining disparity correctness, using correctness predictions to improve disparity estimation, learning the matching function, and end-to-end pipelines.


Early research in stereo using machine learning methodology addressed the problem of deciding whether a disparity was correct or not \cite{cruz95,kong04} but its short-term impact was limited. This changed recently with publications such as the one by Haeusler et al. \cite{haeusler13} who train a random forest to predict the correctness of the output disparities of the SGM algorithm \cite{hirschmuller08} using features computed on the images, disparity maps and matching cost volume.
Gouveia et al. \cite{gouveia_3DV2015} extend the confidence estimator of \cite{spyr14_cvpr} to be applicable to a superpixel-based stereo algorithm. The classifier is able to remove errors from the disparity maps, which are filled in using conventional techniques.
Poggi and Mattoccia \cite{poggi16bmvc} pose confidence estimation as a regression problem and solve it using a CNN trained on small patches of disparity maps based on the observation that patterns in the disparity map can indicate whether a certain disparity assignment is correct. The same authors \cite{poggi17_cvpr} improved a number of previous methods by training a CNN to refine confidence maps.
The classifier's predictions in all cases \cite{gouveia_3DV2015,haeusler13,poggi16bmvc,poggi17_cvpr} are effective in sparsifying the disparity maps by removing potential errors, but do not help in the generation of more accurate disparity estimates.

This shortcoming was addressed by algorithms that inject confidence into the optimization stage.
%
%
Spyropoulos et al. \cite{spyr14_cvpr,spyropoulos_ijcv16} train a random forest on the cost volume to detect ground control points, the disparities of which are favored during MRF-based disparity optimization.
Park and Yoon \cite{park15}  use the predictions of a random forest to modulate the data term of each pixel in SGM-based optimization. Poggi and Mattoccia \cite{poggi16_3dv} present a confidence measure that takes into account multi-scale features and is used to weigh cost aggregation in SGM in order to reduce artifacts. Seki and Pollefeys \cite{seki16,seki17} present two algorithms for adjusting the regularization parameters of SGM using CNNs trained on stereo pairs.


Matching cost estimation was addressed by Li and Huttenlocher \cite{liHuttenlocher08} who use a structured support vector machine to learn linear discriminant functions that compute the data and smoothness terms of a Conditional Random Field (CRF) based on discretized values of the matching cost, image gradients and disparity differences among neighboring pixels. Later, Alahari et al. \cite{alahari10} applied convex optimization, using the same node and edge features as \cite{liHuttenlocher08}, to obtain the solution more efficiently.
Peris et al. \cite{peris12} use synthetic data to train a classifier for matching cost aggregation and disparity optimization. Multi-class LDA is applied to learn a mapping from a feature vector that contains neighborhood matching costs at all disparities for a pixel to the disparity that should be assigned to the pixel.

The approach that ignited the recent wave of deep learning based stereo methods was the one of  \zbontar \cite{zbontar15,zbontar16}. MC-CNN comes in two versions depending on the steps that follow a Siamese network that learns a representation of image patches: in the fast architecture, MC-CNN-fst, the representations of the two patches are compared using the cosine similarity measure, while in the accurate architecture, MC-CNN-acrt, patch similarity is the output of several fully connected layers that operate on the concatenated representations. Similarly to our approach, the networks are trained on matching and mismatching pairs of image patches. \zbontar also augment the training data by distorting and photometrically modifying the images. MC-CNN generates a matching cost volume that undergoes a number of processing steps, including SGM optimization, to generate disparity maps.
Similar Siamese networks followed by the fast or accurate similarity estimation subnetworks have also been proposed by \cite{chen15deep_embed,han2015matchnet,luo_urtasun16,zagoruyko15}, while more recently, other authors have increased the effective receptive field of the networks without loss of resolution \cite{park_lwcnn16,shaked17,ye_ieee17}.
Many of the other top ranked methods have either been inspired by MC-CNN or directly use it to compute the matching cost \cite{barron16,drouyer17,gidaris17,kim16ntde,li_3dmst17,seki16,seki17,taniai17}.

Shaked and Wolf \cite{shaked17} rely on deep learning in two stages of the pipeline: cost computation and final disparity map inference from the matching cost volume. Along with GC-Net \cite{kendall17}, their global disparity network is the only deep learning approach that operates in the cost volume. The network also estimates confidence using a novel reflective loss function.

Disparity refinement is typically addressed by applying various filters and interpolation techniques on the disparity map \cite{mei11}. Recent disparity refinement methods \cite{gidaris17,ye_ieee17} have been able to learn to correct mistakes without relying on hand-crafted rules.

The first end-to-end stereo matching system was introduced by Mayer et al. \cite{mayer16}. The proposed architectures, DispNet and DispNetC, go beyond learning how to match small square patches and learn how to estimate disparity maps given a pair of rectified images.
Kn\"obelreiter et al. \cite{knobelreiter17} present a hybrid CNN-CRF model based on a formulation that allows end-to-end training of a 4-connected CRF, which is more effective on stereo matching than fully-connected CRFs.
%
Very recently, Kendall et al. \cite{kendall17} presented an end-to-end pipeline (GC-Net) based on a high-capacity, deep architecture that resembles the conventional pipeline. It included 3D convolutional layers that regress disparity from a cost volume generated by residual blocks that extract patch representations from the images. Compared to DispNetC, the availability of a cost volume allows GC-Net to exploit context and achieve state-of-the-art results.
Pang et al. \cite{pang17} proposed a cascade of two networks that can be trained end-to-end. The first network is similar to DispNetC while the second refines the disparity map.

\begin{figure*}
  \includegraphics[width=\textwidth]{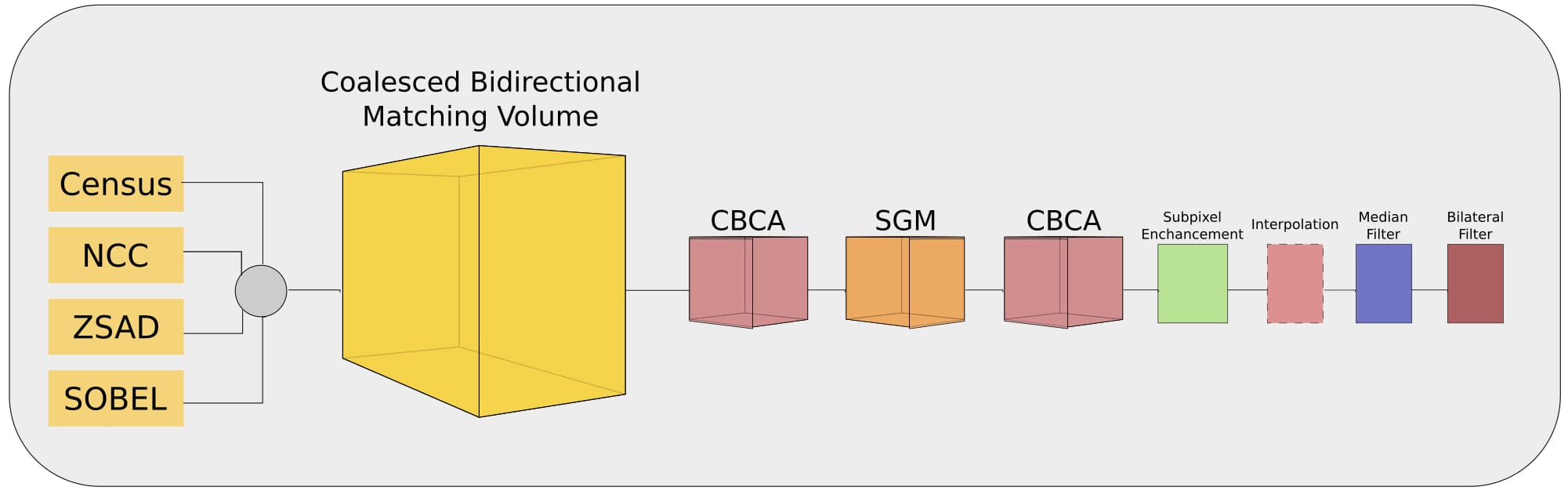}
  \vspace{-14pt}
  \caption{Flowchart of our approach. The matching costs of the four matchers are coalesced with bidirectional confidence features to create the CBMV denoted by the yellow  cube. The smaller cubes show processes that operate on the CBMV, while squares show processes that operate on the disparity map after optimization.}
  \label{fig:flow}
\end{figure*}


%
%


\section{Overview of the approach}\label{sec:overview}

Before presenting a step-by-step break down of our method, we provide a brief high-level description of the building blocks and the intuition behind each step involved. Figure \ref{fig:flow} shows a flowchart of our method.
Our objective is to compute a ``good'' matching volume that captures the support and competition among competing disparity hypotheses and is amenable to global optimization. (See \cite{zbontar16} for an analysis of what a good matching volume is.)

The cost computation step combines the matching volumes computed by four basic matchers
with confidence volumes extracted from the matching volumes.
A Random Forest classifier \cite{breiman01} is trained
to coalesce all the input evidence and generate the CBMV.  The motivation behind this step is to combine the strengths and mitigate the weaknesses of these basic stereo matchers to generate a robust matching volume for the subsequent optimization steps.
Our optimization and post-processing pipeline adopts the steps proposed by \v{Z}bontar and  LeCun with slight modifications to generate the final disparity maps, allowing a direct comparison of our results with those of MC-CNN.
%

\section{Matching Volume Computation}\label{sec:cvc}



The unit on which our algorithm operates is the \emph{matching hypothesis}, ($x_L, x_R, y$), that represents a potential correspondence between pixel $p_L(x_L,y)$ in the left image and a pixel $p_R(x_L-d,y)$ in the right image. Disparity $d$ is always defined as $d=x_L-x_R$ and the matching hypothesis can be written equivalently as ($x_L,d,y$). In the remainder, we drop $y$ for simplicity since the images are rectified. The range of possible disparities $d_{max}$ is also an input.

To determine whether a matching hypothesis is likely or not, we combine \emph{matching volumes} generated by four basic block matching algorithms, \textit{NCC}, \textit{CENSUS},  zero-mean SAD on intensities (\textit{ZSAD})  and SAD on the responses of the horizontal Sobel filter (\textit{SOBEL}), with two \emph{confidence volumes} for each matching function and each matching direction.

\begin{figure}[b]
\begin{center}
\includegraphics[width=\columnwidth]{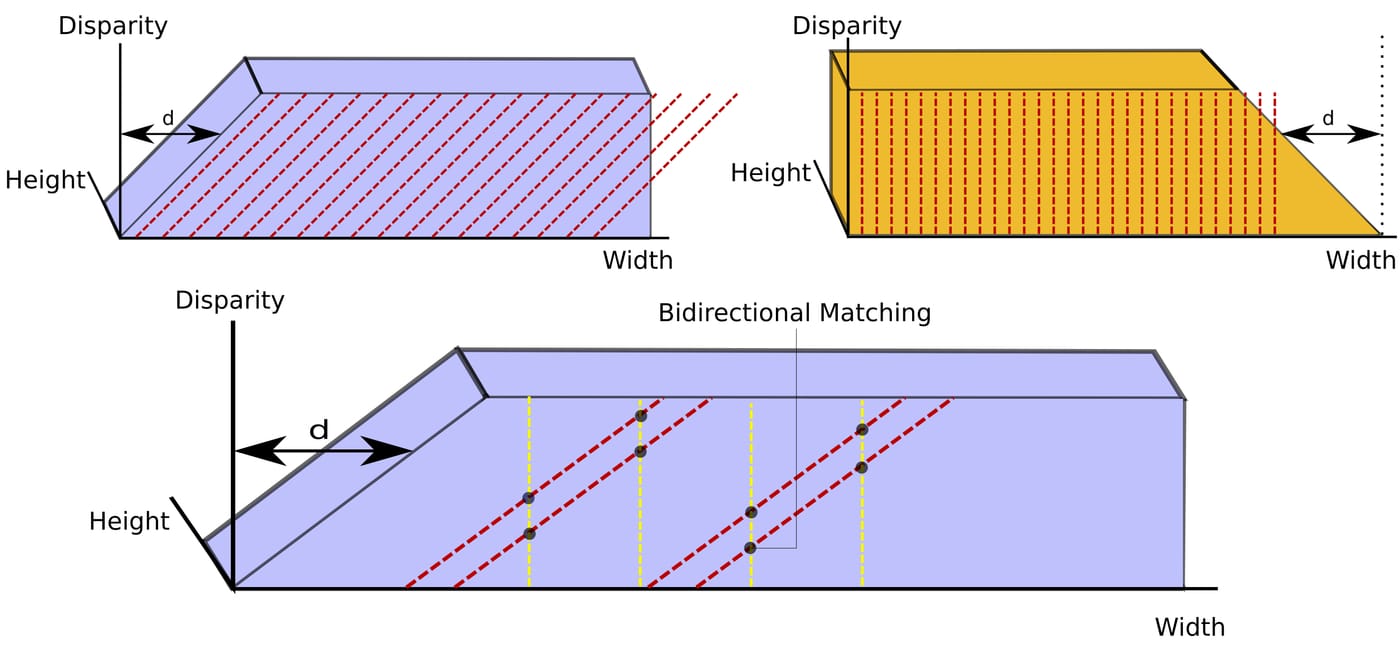}
\end{center}
\vspace{-10pt}
\caption{Top: The left and right matching volumes. The dashed red lines correspond to the matching costs for a given pixel of the right image. A volume can be generated from its counterpart by shifting its equal-disparity slices by $d$. Bottom: For a given element of the matching volume, the ratio and likelihood features are computed along the yellow and red lines corresponding to the right and left epipolar lines respectively. Black dots denote a few intersections of left and right epipolar lines on the matching volume. Each intersection is a matching hypothesis linking a pixel in the left image with a pixel in the right image.  }
\label{fig:costvolumes}
\end{figure}

\pmpar{Matching Volume Representation.} The matching volume for a given matching algorithm stores a value for each possible correspondence between a pixel in the left image to a pixel in the right image within a given disparity range. We use the disparity-based representation for the matching volume and write $C_{cen}(x_L,d)$ for the one computed using \textit{CENSUS} for example.
Given $x_L$ and $d$ (a matching hypothesis), $x_R$ can be retrieved by $x_R=x_L-d$. Typically, the left image is treated as reference and the right image as matching target. Switching the roles of the images, and negating the disparity range, leads to a new matching volume that can also be obtained by re-ordering the values of the left-right volume without re-computation (see Fig. \ref{fig:costvolumes}).

We also compute confidence volumes that capture the ambiguity of disparity hypotheses. These are computed bidirectionally since the competitors of a potential correspondence in the left image are not the same as its competitors in the right image. We introduce the following notation: $c^L_{min}$ denotes the minimum observed cost of a matching algorithm for the left pixel of a matching hypothesis and $d^L_{min}$ the corresponding disparity. $c^R_{min}$ and $d^R_{min}$ are their counterparts for the right pixel. $c^L_{min}$ can be obtained by traversing the yellow lines (constant $x_L$) in Fig. \ref{fig:costvolumes} and $c^R_{min}$ by traversing the red lines (constant $x_R$).

The confidence measures are adapted from \cite{hu12} so that they can be used to compute the confidence of \textit{all} disparity values of a pixel and not only the one with the minimum cost.
For each disparity under consideration and each basic matcher, we extract a feature vector which consists of the following five elements: matching cost $C$, left and right ratio $R^L$ and $R^R$, and left and right likelihood $L^L$ and $L^R$.
We use \textit{CENSUS} as an example below, but the process is repeated for all matchers.

\pmpar{Matching Cost.} This is the raw cost or score of the basic stereo matching algorithm for each disparity under consideration.

\begin{equation}
C_{cen}(x_L,d) = CENSUS(x_L,d)
\end{equation}

\pmpar{Ratio.} The ratio of the minimum cost $c_{{cen,min}}$  over the cost of the candidate disparity $C_{cen}(x_L,d)$ assigns high confidence to disparities with close to minimum cost.

\begin{equation}\label{eq:ratio}
R^L_{cen}(x_L,d)=\frac{c^L_{{cen,min}}}{C_{cen}(x_L,d)}
\end{equation}
\noindent This is computed by finding the minimum cost over $x_L$ along the yellow lines in Fig. \ref{fig:costvolumes}.

Similarly, for the right-to-left matching direction:

\begin{equation}\label{eq:ratio_right}
R^R_{cen}(x_L,d)=\frac{c^R_{{cen,min}}}{C_{cen}(x_L,d)}
\end{equation}
\noindent This is computed by finding the minimum cost over $x_R$ along the red lines in Fig. \ref{fig:costvolumes}.

\pmpar{Likelihood.} Similar to AML in \cite{hu12}, we convert the cost curve for a given pixel to a probability density function to generate a confidence measure for a given disparity hypothesis.

\begin{equation}\label{eq:likelihood}
L^L_{cen}(x_L,d)=\frac{e^{-\frac{(C_{cen}(x_L,d)-c^L_{{cen,min}})^2}{2 \sigma^2}}}{\sum_{i} e^{-\frac{ (C_{cen}(x_L,d_i) - c^L_{{cen,min}})^2}{2 \sigma^2}} }
\end{equation}

\noindent where $\sigma$ is a hyper-parameter that depends on the matching algorithm. To obtain $L^R_{cen}(x_L,d)$ the summation in the denominator must be over $C_{cen}(x_L+d_i,d_i)$ so that all terms match to the same pixel in the right image.

\pmpar{Training.}
We train a Random Forest (RF) classifier to predict the matching likelihood of two image patches.
The classifier learns whether disparity candidates for each pixel are likely to be correct based on the costs, agreements and disagreements of the matchers and the degree of ambiguity along the left and right epipolar lines captured by the confidence measures.

Each disparity hypothesis is represented by 20 cost and confidence values (5 per matching function). Since at most only one hypothesis is correct per pixel, our dataset is imbalanced. To counter this, during training we keep all correct correspondences and sample twice as many incorrect correspondences,
while \zbontar \cite{zbontar16} keep a $1:1$ ratio. We keep all exact correspondences as positive samples, while they also consider correspondences that are off by one disparity level as correct. Then we randomly pick two disparity values, one in the lower range $[0 \dots d_{gt}-1)$ and one in the upper range $(d_{gt}+1 \dots d_{max}]$, where $d_{gt}$ is the ground truth disparity, and label them as negative samples.
The RF learns to make predictions on the correctness of each disparity assignment that links a pair of pixels.

During testing, the RF is applied on the entire matching and confidence volumes to produce the Coalesced Bidirectional Matching Volume, which is the input to the optimization steps described below.

Note that the right CBMV can then be obtained by shifting the left CBMV. The shift is valid under a mild assumption that the left and right confidence values are generated from the same distributions. If this is the case, swapping the left and right confidence features should not affect the classifier's prediction.
That is, $[C \ L^L \ R^L \ L^R \ R^R ]$ and $[C \ L^R \ R^R \ L^L \ R^L ]$ should be equivalent as feature vectors, yielding equal predictions from the classifier.

\section{Optimization and Post-processing}\label{sec:optimization}

The next step of our algorithm is optimization and post-processing to generate the final disparity map from the coalesced volume. Since we need both the left and right disparity maps to apply consistency constraints, we generate the right CBVM by shifting the left one as shown at the top of Fig. \ref{fig:costvolumes}. The following steps are applied to the two volumes separately to produce the two disparity maps.

We use the pipeline of Mei et al. \cite{mei11}, which was also adopted by MC-CNN \cite{zbontar16}. Its steps can be seen on the right side of Fig. \ref{fig:flow}. We only provide a high level description of the components since they are not novel.
There are steps that operate in volumes, namely cross based cost aggregation (CBCA)\cite{zhang09cross}and Semi-Global Matching (SGM) \cite{hirschmuller08}, and further steps that are applied on 2D disparity maps, namely sub-pixel refinement via parabolic fitting, left-right consistency test and interpolation to fill in invalidated pixels, followed by median and bilateral filtering. Following MC-CNN we apply CBCA before and after SGM.
While we keep the structure of the pipeline, we tuned the values of its parameters via cross-validation per dataset to obtain high accuracy. More details are included in the supplement.

\section{Experimental Results}\label{sec:experiments}

\begin{table*}[t]
	\small
	\begin{center}
		\scalebox{0.92}{
			\begin{tabularx}{\textwidth}{lXXXXXXXX}
				\hline
				&\multicolumn{4}{c}{Out-Noc} &\multicolumn{4}{c}{Out-all}\\
				\hline
				Method		&bad $.5$			&\textbf{bad $1.0$}	 &bad $2.0$  &rms	&bad $.5$	&bad $1.0$	&bad $2.0$	&rms\\
				\hline													
				CBMV		&\textbf{13.69\%}	&\textbf{5.35\%}	&\textbf{1.56\%} &\textbf{0.71}	&\textbf{14.52\%}	&\textbf{5.97\%} 	&\textbf{1.97\%}	&\textbf{0.86}\\
				SGM\_ROB	&19.52\%			&10.08\%			&4.07\%			&1.89			&20.33\%			&10.77\% 	&4.67\%	&2.11\\
				MeshStereo	&22.27\%			&11.52\%  			&5.78\%			&1.21			&22.95\%			&11.94\%	&6.09\%	&1.29\\
				SPS-STEREO	&55.62\%			&15.04\%			&3.08\%			&1.07			&56.02\%			&15.83\%	&3.67\%	&1.22\\
				SGM-STEREO 	&54.67\%			&15.62\%  			&4.39\%			&1.83			&55.54\%			&17.25\%	&6.27\%	&2.67\\
				ELAS		&33.66\%			&16.72\%			&8.05\%			&1.89			&34.78\%			&17.99\%	&9.07\%	&1.52\\
				\hline	
		\end{tabularx}}
	\end{center}
	\caption{Results of our method (CBMV) on ETH3D two-view benchmark. Our method outperforms all other methods by a large margin. All our submissions, including on ETH3D, use the same model, trained on the Middlebury 2014 dataset. The methods are sorted based on the main validation metric: bad $1.0$ out-noc.   }
	\label{tab:eth3d}
\end{table*}

We evaluate our algorithm on the 2014 version of the Middlebury Stereo Evaluation dataset \cite{scharstein14}, the 2012 and 2015 versions of the KITTI stereo benchmark \cite{geiger13,Menze2015CVPR} and the ETH3D stereo benchmark \cite{schoeps2017cvpr}.

The Middlebury dataset consists of a training set of 15 stereo pairs, 13 additional stereo pairs, all with publicly available ground truth, and a test set of 15 stereo pairs, the ground truth for which has not been released. Compared to previous versions of the benchmark, this version is more challenging because most stereo pairs have imperfect rectification, except those with a suffix 'P' in their filename, while several others contain images taken under different exposure or lighting, denoted by 'E' and 'L' respectively \cite{scharstein14}. The image resolution varies between 1.5 and 5.9 megapixels with an average of 5.2 megapixels and the disparity range varies between 256 and 800. 
As most authors, we use half-resolution images. 
The ranking in the new tables is determined by weighted averages of the selected metric.

The KITTI datasets consist of approximately 200 training and 200 testing stereo pairs each. The ground truth is semi-dense covering roughly 30\% of all pixels and is concentrated in the lower part of the images. The ground truth of the test sets has been withheld. An important difference between the two versions of the benchmark is that cars have been manually annotated in the 2015 version and have dense ground truth, including their windshields. The latter are explicitly deleted from the ground truth of the 2012 benchmark.

The ETH3D stereo dataset consists of 27 training and 20 testing stereo pairs. The ground truth is dense and generated by a Faro Focus X 330 laser scanner. The ground truth for the training set is publicly available,while for the test set, it has not been released. 

\paragraph{Experimental Setup.}
To compute the initial matching volumes on the Middlebury data using the four block matching algorithms, we set the width of the matching windows to: $3 \times 3$ for \textit{NCC}, $5 \times 5$ for \textit{ZSAD}, $11 \times 11$ for \textit{CENSUS} and $5 \times 5$ for \textit{SOBEL}. The $\sigma$ parameter in Eq. (\ref{eq:likelihood}) was set to 0.02 for \textit{NCC}, 100 for \textit{ZSAD} and \textit{SOBEL} and 8 for \textit{CENSUS}. Parameters for the KITTI data are similar and are included as supplementary material.

On the Middlebury 2014 dataset, we use three-fold cross-validation to train our RF classifier. We split the training set into three sets of five images. Two of these sets of five and the set of 13 additional images, which are available with the dataset but are not evaluated, are used for training during each fold of the cross-validation process, while the remaining five images are used for testing. Before the final testing phase on the Middlebury test set, we train our classifier on all 28 training stereo pairs.
Due to the availability of more data, we use two-fold cross validation on the KITTI datasets. We did not train on the ETH3D dataset.
%

We tune the optimization parameters (see Section \ref{sec:optimization}) separately for each dataset, as in \cite{zbontar16}. A complete hyper-parameter configuration is provided in the supplement.

\begin{table}[h]
\small
\begin{center}
\scalebox{.92}{
\begin{tabularx}{\linewidth}{lXXXX}
\hline
 			&{\fontsize{8}{10}\selectfont bad-2.0 nonocc}			&{\fontsize{8}{10}\selectfont bad-2.0 \ \ \ \ \ \  all} 		 &{\fontsize{8}{10}\selectfont avgerror all}		&{\fontsize{8}{10}\selectfont rms-error all} \\
\hline
\multicolumn{5}{l}{\fontsize{8}{10}\selectfont Middlebury 2014 test set}\\
\hline					
MC-CNN-acrt	&\textbf{8.08}\%	&\textbf{19.1}\%	&$17.9$	&$55.0$	\\
CBMV(ours)	&$11.1$\%	&$20.5$\%	&$\textbf{14.4}$	&\textbf{46.9}	\\
MC-CNN-fst	&$9.47$\%	&$20.6$\%	&$19.3$	&$55.7$	\\
\hline					
\multicolumn{5}{l}{\fontsize{8}{10}\selectfont Middlebury 2014 training set}\\					
\hline					
MC-CNN-acrt	&\textbf{10.1}\%	&\textbf{19.7}\%	&$11.8$	&$36.6$	\\
CBMV(ours)	&$11.7$\%	&$20.3$\%	&\textbf{11.5}	&\textbf{34.9}	\\
MC-CNN-fst	&$11.7$\%	&$21.5$\%	&$12.8$	&$37.5$	\\
\hline
\end{tabularx}}
\end{center}
\caption{Results of our method (CBMV) on the Middlebury 2014 test and training sets, compared with the results of MC-CNN-acrt and MC-CNN-fst using four different metrics.}
\label{tab:mvs_comparison_train}
\end{table}

\begin{figure*}
\begin{center}
\begin{tabular}{cccccc}
\multicolumn{2}{c}{Middlebury model} & \multicolumn{2}{c}{KITTI 2012 model} & \multicolumn{2}{c}{KITTI 2015 model} \\
\includegraphics[width=\widh]{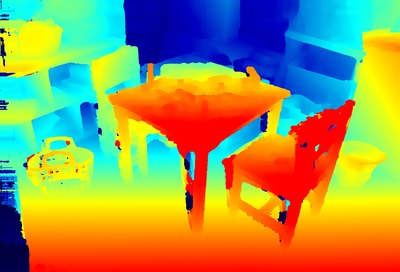} & \includegraphics[width=\widh]{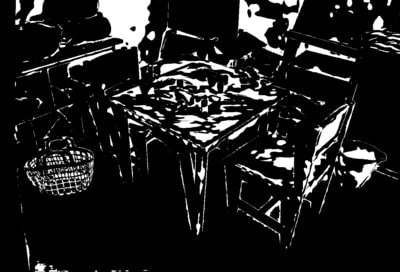}  & \includegraphics[width=\widh]{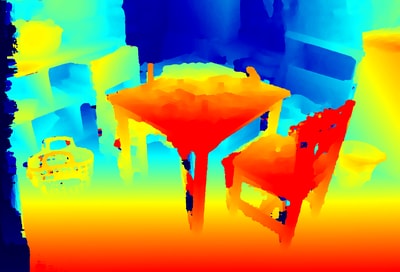} & \includegraphics[width=\widh]{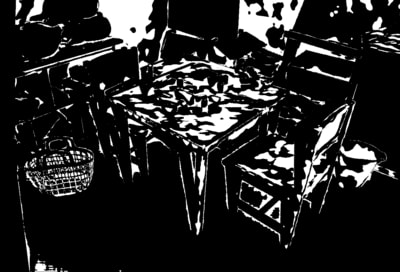} & \includegraphics[width=\widh]{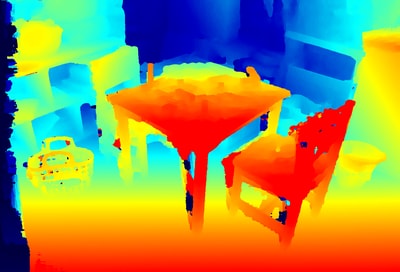} & \includegraphics[width=\widh]{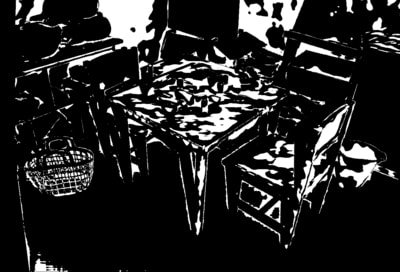}\\
\multicolumn{2}{c}{Error 13.3\%} & \multicolumn{2}{c}{Error 15.86\%} & \multicolumn{2}{c}{Error 15.92\%} \\
\end{tabular}
\end{center}
\vspace{-5px}
\caption{Generalization examples of CBMV on the Playtable data from Middlebury. Our method is robust when tested in different domains, especially compared to MC-CNN. The corresponding error rates at a 2-disparity level tolerance for MC-CNN-fst are 18.0\% , 41.43\% and 38.67\% respectively.   }
\label{fig:dom_fig}
\end{figure*}

%
%
%
%

\begin{table*}[h]
	\small
	\begin{center}
\vspace{-8pt}
		\scalebox{.92}{
			\begin{tabularx}{\linewidth}{lXXXXXXXXXXX}
				
				\hline
				&	&	&	\multicolumn{9}{c}{Test set}\\
				&	&	&\multicolumn{3}{c}{KITTI 2012 (Out-Noc) } & \multicolumn{3}{c}{KITTI 2015 (Out-All)} & \multicolumn{3}{c}{Middlebury (bad 2.0) }\\
				\hline
				&	& & MC-ac	& MC-fst & CBMV & MC-ac &  MC-fst & CBMV & MC-ac & MC-fst & CBMV \\	
				
			\multirow{3}{50pt}{Training set}	& \multicolumn{2}{r}{KITTI 2012}	& 0\% & 0\%  & 0\% 	& 23.07\% & 13.28\% & -0.41\%   	&40.20\%  &33.58\% & 7.00\%\\
				
				& \multicolumn{2}{r}{KITTI 2015}	& 63.98\% & 17.54\% & 3.02\% 	& 0\% & 0\% & 0\%   				&79.39\% &41.62\% & 7.69\% \\
				
				& \multicolumn{2}{r}{Middlebury}   & 17.62\%  & 10.51\% & -4.62\% 	& 38.15\%	& 18.79\% & -2.09\%	 	& 0\%	& 0\%  & 0\% \\				
				\hline
		\end{tabularx}}
	\end{center}
	\vspace{-5pt}
	\caption{Quantitative generalization results for CBMV, MC-CNN-fst and MC-CNN-acrt.  This table shows relative increases in error rate when the training set is different than the test set. For example, the rightmost column means that the CBMV error rate increases by 7\% when trained on KITTI 2012 and tested on Middlebury, compared to training and testing on Middlebury.
%
For MC-CNN-acrt we use data from Table 10 of \cite{zbontar16}. }
	\label{tab:domain_adapt}
\end{table*}

\begin{figure*}
\begin{center}
\begin{tabular}{ccc}
\multicolumn{3}{c}{Middlebury 2014}\\
\includegraphics[width=\wid]{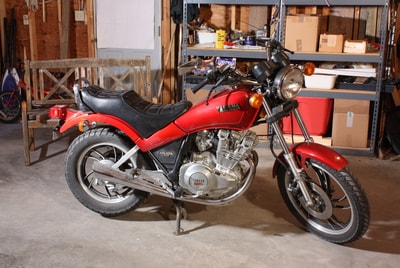} & \includegraphics[width=\wid]{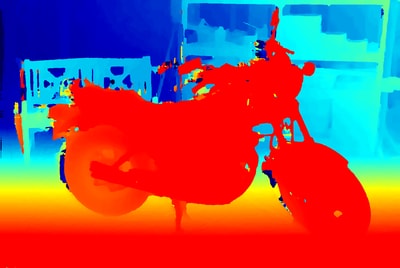}  & \includegraphics[width=\wid]{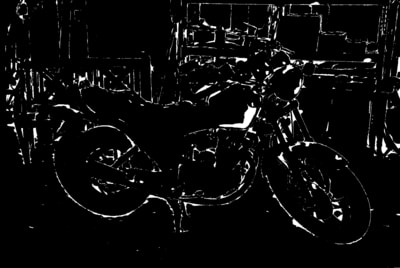}\\
\includegraphics[width=\wid]{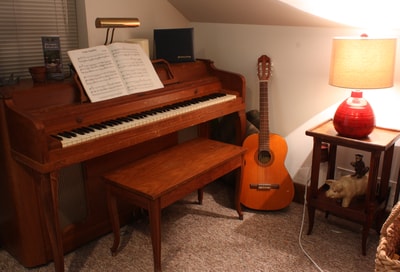} &\includegraphics[width=\wid]{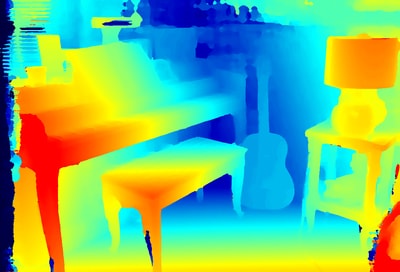}&\includegraphics[width=\wid]{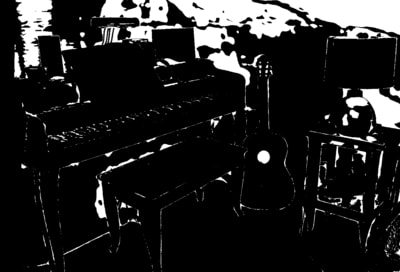} \\
\multicolumn{3}{c}{KITTI 2012}\\
\includegraphics[width=\wid]{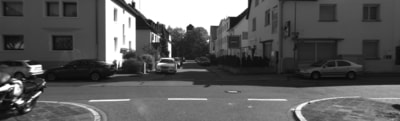} &\includegraphics[width=\wid]{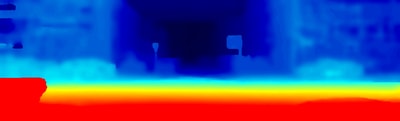}&\includegraphics[width=\wid]{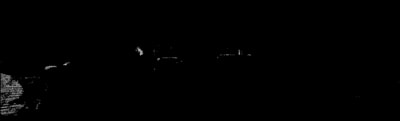} \\
\includegraphics[width=\wid]{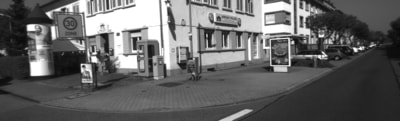} &\includegraphics[width=\wid]{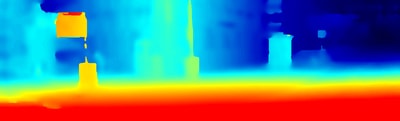}&\includegraphics[width=\wid]{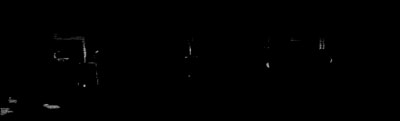} \\
\multicolumn{3}{c}{KITTI 2015}\\
\includegraphics[width=\wid]{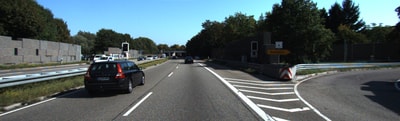} &\includegraphics[width=\wid]{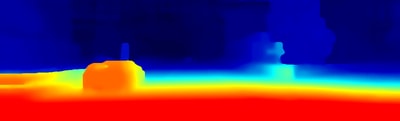}&\includegraphics[width=\wid]{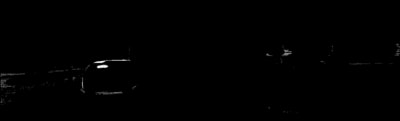} \\
\includegraphics[width=\wid]{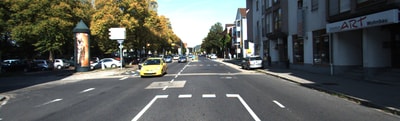} &\includegraphics[width=\wid]{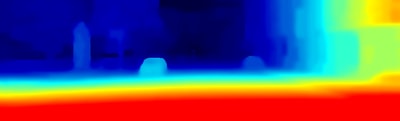}&\includegraphics[width=\wid]{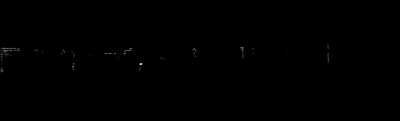} \\

Input Image & CBMV & Error map \\
\end{tabular}
\end{center}
\caption{Results of our method (CBMV) on the three datasets: Middlebury 2014, KITTI 2012 and KITTI 2015. Note that the KITTI disparity maps were generated using an RF trained on Middlebury 2014.  }
\label{fig:MVS_RF}
\end{figure*}

\paragraph{Results.}
On the Middlebury data, the final disparity maps computed by CBMV have an average error rate of $11.1$\% and $11.7$\% out of non-occluded pixels on the testing and training set, respectively, with the error tolerance set to 2.0 disparity levels. These results show that our method can produce competitive results with the MC-CNN-fst architecture. However, our method outperforms MC-CNN-fst on both training and testing sets when we consider all pixels and shows competitive results to the MC-CNN-acrt architecture. Moreover our method ranks higher than both MC-CNN-fst and MC-CNN-acrt with respect to average and RMS error. Table \ref{tab:mvs_comparison_train} shows a comparison of our method with MC-CNN-acrt and MC-CNN-fst. Figure  \ref{fig:MVS_RF} shows disparity maps generated by our method with corresponding error maps on the Middlebury 2014, KITTI 2012 and KITTI 2015 datasets.



\pmpar{Generalization.}\label{dom_adapt}
Being able to evaluate a method on the available benchmarks is very important since a it allows fair, comprehensive comparisons with other methods. However, benchmarks cannot capture all the difficulties involved when deploying a method on the field. To evaluate the transferability of our method, we use our trained RFs on the three different datasets, Middlebury 2014, KITTI 2012 and KITTI 2015, and test on the corresponding unseen training sets. As an example, we use the RFs from KITTI 2012 and KITTI 2015 to test on the Middlebury 2014 training set. The same approach is used for the downloaded MC-CNN-fst models, which serve as baselines.
Since the optimization and post-processing stage is an integral part of both methods, for fairness and to avoid inconclusive results due to hyper-parameter tuning, we keep the hyper parameter values that worked best when the method is applied to the same dataset. More precisely when we use the RF trained on KITTI 2012 to test on the Middlebury 2014 training set, we use the Middlebury 2014 best hyper-parameters during testing. We were unable to run MC-CNN-acrt KITTI models on the Middlebury dataset due to the limited amount of global memory on the GPU, but we include results based on the numbers reported in \cite{zbontar16}. It is worth noting that MC-CNN-acrt is significantly worse that MC-CNN-fst in this particular experiment, which shows that MC-CNN-acrt specializes even more to the particular training dataset to achieve higher accuracy.

Table \ref{tab:domain_adapt} shows that our method adapts much better to new unseen environments. Figure \ref{fig:dom_fig} shows results on a particular example where MC-CNN-fst has a dramatic drop in accuracy. In Table \ref{tab:domain_adapt} we can see that our Middlebury RF outperforms the RFs trained on KITTI 2012 and 2015. We believe that this behavior can be attributed to that fact that Middlebury 2014 dataset contains much more diverse scenes, thus the classifier can generalize better.

We submitted results to the KITTI 2012, KITTI 2015 and ETH3D test sets using the RF trained on the Middlebury 2014 dataset and optimization and post-processing parameters tuned on each target dataset. The error rates of our disparity maps are  $3.56\%$  and $4.73\%$, on non-occluded and on all pixels on KITTI 2012, and $4.58\%$ and $5.06\%$ on non-occluded and on all pixels on KITTI 2015. Table \ref{tab:eth3d} shows a comparison of our method with other methods on the ETH3D benchmark. Our method outperforms every other method by a large margin. We are not aware of any other submission that was not trained on data from KITTI or ETH3D itself. Please visit the KITTI and ETH3D websites for comparisons with other methods.

\pmpar{Runtime.}
At first glance our method seems very expensive both in computation and space requirements. This is partially true. Computing the four initial cost volumes is very efficient and can be done in parallel. In our implementation the total time spend computing the four initial costs is approximately 2.5 seconds for a KITTI stereo pair. The feature extraction process takes about 7 seconds. To lower the space requirements, features can be extracted in batches of image rows. The bottleneck of our method is the random forest classifier which takes approximately 162 seconds. A better implementation of the random forest where training is done on the CPU and inference is performed on the GPU is feasible but out of the scope of the current paper. Moreover, due to the robustness of our model, an ASIC RF implementation is possible and would enable very high frame rates. Most optimization and post-processing steps have to be executed for both the left and right disparity map, but they only take a few seconds. The total runtime of our method on KITTI is 250 seconds. Our complete implementation including the trained RF model is available at \url{https://github.com/kbatsos/CBMV}.

\section{Conclusions}\label{sec:conclusions}

We have proposed a novel approach for estimating a bidirectional matching volume by coalescing matching and confidence data generated by applying conventional matching functions on rectified stereo pairs. We have evaluated the accuracy and the generalizability of this approach quantitatively and qualitatively.

Comparing the results of CBMV with those of MC-CNN on the 2014 Middlebury benchmark, we observe that CBMV is superior with respect to average and RMS disparity errors when all pixels are considered. Considering other error metrics on both non-occluded and all pixels, the ordering of the two MC-CNN architectures and CBMV fluctuates. It would be fair to say that MC-CNN-acrt is the most accurate overall, with the other two methods being essentially tied.


The advantage of our method lies in its generalizability. According to Tonioni et al. \cite{tonioni17}, end-to-end deep architectures \cite{mayer16} tend to specialize on their training domain. MC-CNN is better suited for previously unseen domains, but as we have shown in Table \ref{tab:domain_adapt}, our method generalizes much better. Training on the Middlebury data resulted in even higher accuracy on the KITTI benchmark than training on the target dataset itself. Transferring RFs in the other direction led to a small loss of accuracy. 
%
\emph{The strength of our approach is that, due to its design that avoids learning directly from image appearance, trained classifiers can be applied in domains without ground truth data.} This is a critical feature for being able to apply a method on images taken in the field (not necessarily in real time).

In our future work we plan to investigate ways of combining the generalization properties of our approach with the advantages of end-to-end deep learning architectures.

\pmpar{Acknowledgments.} Research reported in this publication was supported by the National Science Foundation under Awards IIS-1527294 and  IIS-1637761. We are grateful to Enrique Dunn for constructive discussions, especially on the name of our approach.

\newpage

{\small
\bibliographystyle{ieee}
\bibliography{learningStereo}
}

\end{document}